# The Signal in the Noise

## An Auditable Reliability Layer for Biomedical Text Classification[1]

Moustafa Yehia Hassan
Doha Institute for Graduate Studies (DI)
Asia Pacific University of Technology & Innovation (APU)
Doha, Qatar; Kuala Lumpur, Malaysia
moustafa.hassan@dohainstitute.edu.qa

Sharon Wong
Asia Pacific University of Technology & Innovation (APU)
Kuala Lumpur, Malaysia
TP087393@mail.apu.edu.my
Woh Kai Xuan
Asia Pacific University of Technology & Innovation (APU)
Kuala Lumpur, Malaysia
TP056443@mail.apu.edu.my

**Abstract- Biomedical NLP pipelines routinely presuppose clean input text, yet large-scale corpora assembled through automated PDF parsing harbour pervasive OCR-like artifacts, token splits and merges, hyphenation remnants, and character-level corruption, that systematically erode lexical evidence and degrade downstream classifiers. We introduce a conservative, fully auditable spell-correction reliability layer conceived as a safety-oriented preprocessing module rather than a maximal-accuracy corrector: under conditions of uncertainty, the system abstains from editing, in accordance with a medical do-no-harm philosophy. The deterministic architecture couples bounded edit-distance candidate generation with corpus-derived n-gram scoring and a suite of biomedical safety gates that protect domain-critical terminology. We evaluate the layer both intrinsically, on a manually curated benchmark of 2,104 token-level cases, and extrinsically, on a tri-class CORD-19 topic classifier (Prevention, Treatment, Epidemiology) spanning 10,000 examples under a principled four-run protocol (Clean, Noisy, Restored, Safety). Intrinsically, the layer attains 94.61% error-fix recall on synthetic errors with zero harmful edits on negative controls. Downstream, it recovers approximately 80.45% of the noise-induced macro-F1 degradation, elevating macro-F1 from 0.7654 (Noisy) to 0.7717 (Restored) while preserving near-clean performance (Safety: 0.7721). A supplementary case study on 103 real-world OCR-extracted abstracts classified with BioBERT confirms that transformer encoders appeared relatively robust to mild noise, motivating a future grey-box architecture that integrates bounded neural signals and UMLS lexicons without compromising auditability. The system is fully deterministic, artifact-driven, and designed with deployment and auditability in mind.**

## I. INTRODUCTION

Biomedical NLP systems are routinely developed under the tacit assumption that input text is well-formed. In practice, however, large-scale corpora derived from automated PDF ingestion and OCR harbour systematic extraction artifacts that corrupt lexical evidence. The consequences are especially severe in the biomedical domain, where critical discriminative signals reside in sparse, long-tail terminology [1]. While spell correction is an intuitive remedy, indiscriminate editing risks silently rewriting valid domain entities into common-language surrogates, causing high-impact semantic distortions. Because a single harmful edit can outweigh numerous benign fixes, we frame this as an asymmetric-risk problem. We therefore introduce an upstream reliability layer focused on safe robustness, explicitly favoring abstention under uncertainty in accordance with the medical do-no-harm imperative [11].

Our principal contributions are as follows:

- A conservative, fully white-box reliability layer that performs bounded spell correction with explicit margin-based abstention and biomedical safety gates, ensuring that no edit is applied without traceable evidential justification.
- A deterministic, reproducible implementation that persists all learned artifacts (vocabulary, n-gram counts, configuration) and emits comprehensive token-level decision traces for every correction considered.
- A principled four-run evaluation protocol (Clean, Noisy, Restored, Safety) that jointly quantifies robustness recovery under noise and potential harm on pristine inputs.

 Accepted for publication in the 2026 14th International Conference on Bioinformatics and Computational Biology (ICBCB 2026), Kitakyushu, Japan; final version to appear in IEEE Xplore.

## II. RELATED WORK

### A. Spell correction

Classical spell correction is conventionally decomposed into (i) candidate generation within a bounded edit radius using efficient indexing structures [3, 4] and (ii) candidate ranking via noisy-channel models or language-model evidence [6]. Delete-based indexing (e.g., SymSpell [5]) renders candidate retrieval computationally efficient for edit distances ≤2, while lightweight n-gram scoring provides auditable contextual disambiguation. In the biomedical domain, the persistent tension between recall on true errors and protection against overcorrection is exacerbated by long-tailed vocabularies rich in abbreviations, gene symbols, and chemical identifiers [7, 8]. Our work foregrounds the deployment objective: safe robustness under realistic extraction noise, rather than maximal correction coverage.

### B. Extraction noise and robustness

OCR and PDF-extraction artifacts are well-documented sources of NLP performance degradation [9]. Synthetic corruption is frequently employed to approximate such noise under controlled experimental conditions [14]. Although neural rewriting and LLM-based cleanup can attenuate visible errors, these approaches are often non-deterministic and resist token-level tracing, posing challenges in regulated biomedical settings where preprocessing must be independently auditable and reproducible across environments.

### C. Trustworthy biomedical NLP

In healthcare-adjacent NLP, trustworthy AI mandates transparency, traceability, and safe failure modes [11]. Abstention, preserving the original input when evidential support is insufficient, constitutes a direct mechanism for safe failure. We operationalize this principle within the correction domain and measure its downstream impact through an explicit clean-text safety evaluation.

## III. METHODOLOGY

Given a token sequence T = ($t_1$, …, $t_n$) extracted from a biomedical title+abstract, potentially corrupted by OCR-like noise, the reliability layer proposes local substitutions for tokens suspected of corruption, but commits an edit exclusively when evidential support is strong. The architecture is intentionally white-box: candidate generation employs bounded edit distance (≤2), candidates are ranked by corpus-derived unigram and bigram log-probabilities with add-k smoothing, and all decisions are constrained by explicit safety gates and margin-based abstention thresholds.

Vocabulary and n-gram artifacts. From the filtered corpus we construct: (i) a vocabulary V of observed tokens with frequency counts, and (ii) bigram counts for adjacent token pairs. Probabilities are estimated with add-k smoothing:

$$P_{uni}(w) = \frac{\mathrm{count}(w) + k}{\sum_{u \in V} \mathrm{count}(u) + k|V|}$$

$$P_{bi}(u, w) = \frac{\mathrm{count}(u, w) + k}{\mathrm{count}(u) + k|V|}$$

All artifacts are persisted and versioned so that the same correction behaviour is reproducible across runs and environments. Each candidate c is scored as:

$$s(c) = \lambda_1 \log P_{uni}(c) + \lambda_2 \log P_{bi}(p, c) + \lambda_3 \log P_{bi}(c, n) - \alpha \cdot \mathrm{ED}(t, c) \quad (1)$$

where p and n denote the preceding and following tokens, λ weights balance unigram and bigram evidence, and α penalizes larger edit distances. We bias toward the original token by requiring a margin:

$$s(c^*) - s(t) \geq \delta \quad (2)$$

so that small score fluctuations do not trigger edits. This "original-bias" is a key reliability property: if the system is uncertain, it preserves the text rather than introduce an untraceable mutation.

*Fig. 1*

*The proposed reliability layer pipeline architecture*

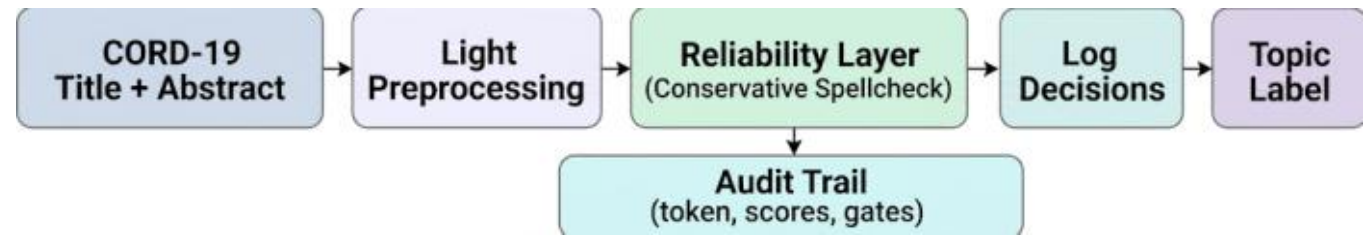

**Algorithm 1** Conservative token correction with abstention

```
Require: token t, context (p, n), vocab V, n-grams NG
NG
 1: if Protected(t) then
 2:    return t
 3: end if
 4: Cand ← Candidates(t, V, ED ≤ 2)
 5: best ← t; bestScore ← Score(t)
 6: for c ∈ Cand do
 7:    s ← Score(c | p, n) −  α · ED(t, c)
 8:    if s > bestScore then best ← c;  bestScore ← s
 9: end for
10: if best = t then return t
11: if RealWord(t) and bestScore − Score(t) < δ_rw then return t
12: if bestScore − Score(t) < δ then return t
13: return best
14: return best
15: return best − Score(t) < δ
13: return best
```

Candidate generation. For each token t absent from the in-domain vocabulary V, we retrieve all candidates within edit distance ≤2 via a delete-index structure [5]. Distance 1 captures the majority of single-character OCR errors; distance 2 accommodates frequent split/merge artifacts while maintaining tractable candidate sets. Low-frequency candidates are pruned, and repeated noisy strings are cached to minimize overhead in batch pipelines.

Scoring and abstention. Candidates are scored using a composite function incorporating unigram frequency and left/right bigram context. The system edits only when the best candidate's score exceeds the original token's score by a margin ≥ δ, calibrated on held-out development data. For tokens already present in the vocabulary V (real-word cases), a stricter margin δ_rw is enforced. This "original-bias" principle is a central reliability property: when the system is uncertain, it preserves the text rather than introducing an untraceable mutation.

Biomedical safety gates. To minimize the risk of harmful rewrites, multiple deterministic gates short-circuit correction for high-risk patterns: a short-token gate protects tokens of length ≤3; a pattern gate preserves alphanumeric biomedical identifiers, all-caps abbreviations, and numeric-heavy tokens; and a real-word gate enforces a stricter margin δ_rw for in-vocabulary tokens. Each gate activation is a transparent, auditable rule recorded in the decision trace.

*TABLE I*

SAFETY GATES (DESIGN INTENT).

| Gate | Purpose |
|---|---|
| Short-token | Avoid unstable edits on low-information tokens |
| Pattern-based | Preserve biomedical IDs and symbol-rich tokens |
| Real-word | Prevent rewriting valid rare biomedical tokens |
| Score margin | Prefer “no change” unless evidence is strong |

Audit trail. For every correction candidate considered, the system persists a comprehensive token-level trace comprising: the original token and its positional context, the complete candidate list with associated scores, the computed margin and applicable threshold, activated safety gates, and the final decision (apply or abstain). This facilitates rigorous post-hoc inspection, error analysis, and exact reproducibility across environments.

*TABLE II*

MINIMAL AUDIT-TRACE FIELDS (TOKEN-LEVEL).

| Field | Meaning |
|---|---|
| token, position | original string and index |
| protected, gate | whether a safety gate fired |
| candidates | candidate list (top-$K$) |
| scores | contextual scores per candidate |
| margin | $s(c^*) - s(t)$ |
| decision | apply / abstain |

# IV. IMPLEMENTATION AND REPRODUCIBILITY

A central engineering choice is the clear boundary between learned artifacts (vocabulary counts, bigram counts, configuration values) and runtime logic (tokenization, gates, scoring, trace emission). This separation ensures reproducibility: given the same artifacts, identical text produces identical corrections and audit traces. Parameters {λ1, λ2, λ3, α, δ, δrw} are selected on a held-out split via constrained grid search: maximize recall and downstream recovery subject to zero harmful edits on negative cases. In deployment, we recommend treating correction as a switchable module: (i) enable it only for sources with extraction noise, (ii) log per-document edit rates, (iii) alert on abnormal rates as upstream parser failures, and (iv) retain raw text for traceability. An audit UI enables a closed-loop workflow, observe error buckets, adjust rules, rebuild artifacts, re-run the four-run protocol, emphasizing transparent iteration over black-box re-training.

# V. EXPERIMENTAL SETUP

Data. We draw upon CORD-19 [1] title+abstract pairs mapped to a tri-class topic task (Prevention, Treatment, Epidemiology) under a strict weak-supervision labeling policy: weighted keyword evidence determines class assignment, with conservative handling of ties and low-evidence cases, which are excluded rather than forced [10]. This yields scalable silver labels suitable for robustness evaluation while maintaining interpretable label-assignment rules. Crucially, we do not claim these labels are equivalent to expert ground truth; they serve as a pragmatic mechanism for studying robustness behavior under controlled noise. For intrinsic evaluation, we manually curated and annotated 2,104 token-level cases spanning synthetic OCR-like corruptions and negative controls.

*TABLE III*

POSITIONING RELATIVE TO COMMON APPROACHES (QUALITATIVE).

| Approach | White-box | Deterministic | Abstains | Token-level audit | Deployment fit |
|---|---|---|---|---|---|
| Edit-distance + LM [6] | Yes | Yes | Usually no | Partial | Medium |
| SymSpell variants [5] | Yes | Yes | No | Low | Medium |
| LLM rewriting | No | Often no | No | Low | Variable |
| Weak supervision (labels) [10] | Partial | Yes | N/A | Medium | High |
| **Ours (reliability layer)** | **Yes** | **Yes** | **Yes** | **High** | **High** |

Corruption model. To rigorously stress-test robustness, we inject synthetic OCR-like artifacts emulating common PDF/OCR failures: character substitutions (weight 0.50), deletions (0.25), and insertions (0.25), applied with length-aware probability:

$$P(\text{error}|t) = 0.15 \cdot |t|^{-0.5} \quad (3)$$

This model is not claimed to perfectly replicate real PDF parsers; it serves as a controlled stress test producing consistent “Noisy” distributions across experimental repeats.

Four-run protocol and metrics. We train a lexical classifier (TF–IDF + logistic regression with class-weight balancing) on clean text and evaluate under four conditions: (i) Clean (original input), (ii) Noisy (corrupted input), (iii) Restored (noisy text corrected by the layer), and (iv) Safety (clean text passed through the layer).

We report the macro-averaged F1 score alongside the Error Reduction Rate (ERR), which quantifies the fraction of noise-induced performance degradation recovered by the reliability layer. The ERR is formulated as:

$$\text{ERR} = \frac{\text{F1(Restored)} - \text{F1(Noisy)}}{\text{F1(Clean)} - \text{F1(Noisy)}}$$

Furthermore, we establish the statistical stability of the observed recovery by computing bootstrap 95% confidence intervals for ΔF1 over 5,000 resampling iterations.

# VI. RESULTS

## A. *Intrinsic correction quality*

On the 2,104-case intrinsic benchmark, the layer attains 94.61% recall on synthetic errors while producing zero harmful edits on negative controls, a result that directly operationalizes the intended conservative operating point wherein safety takes precedence over maximal recall (Table IV).

TABLE IV

INTRINSIC EVALUATION ON CURATED BENCHMARK (2,104 CASES).

| Metric | Value |
|---|---|
| Error-fix recall (true errors) | 94.61% |
| Harmful edits on negatives | 0 |

*B. Downstream topic classification*

Under synthetic corruption, macro-F1 drops from 0.7732 (Clean) to 0.7654 (Noisy), representing a degradation of 0.0078. Application of the reliability layer recovers the majority of this loss: macro-F1 rises to 0.7717 (Restored), yielding ERR ≈ 80.45%. Passing clean text through the layer preserves near-pristine performance (Safety: 0.7721; safety drop = 0.0011). Bootstrap 95% confidence intervals for ΔF1 are strictly positive [0.0001, 0.0122], confirming that the observed recovery is statistically robust and consistent with a non-zero recovery under bootstrap resampling (Table V; Fig. 2).

TABLE V:

DOWNSTREAM MACRO-F1 UNDER THE FOUR-RUN PROTOCOL (10,000 EXAMPLES).

| Run | Macro-F1 |
|---|---|
| Clean | 0.7732 |
| Noisy | 0.7654 |
| Restored | 0.7717 |
| Safety | 0.7721 |

Fig. 2

*Macro-F1 scores across the four-run evaluation protocol.*

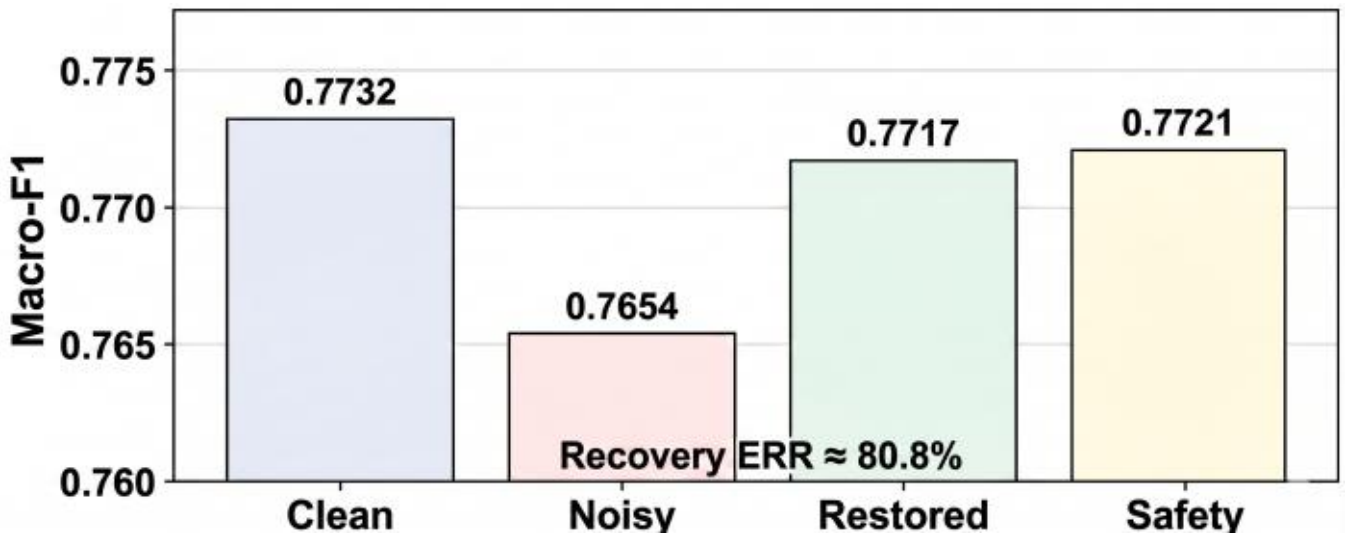

*C. Qualitative correction examples*

Representative cases illustrate the layer's decision-making: "remdesivlr" → "remdesivir" (strong unigram + context); "C0vid-l9" → "COVID-19" (pattern-gate permits, strong evidence); "IL-6" → abstain (biomedical ID protected); "hydroxy-chloro-quine" → "hydroxychloroquine" (split repaired with context). These demonstrate the asymmetric-risk design: high-confidence corrections proceed while ambiguous cases default to safe abstention.

## VII. ABLATIONS

Systematic ablation analysis confirms that both contextual bigram evidence and margin-based abstention are indispensable components. Removing bigram context elevates the rate of false-positive edits, while relaxing the abstention threshold improves recall at the cost of occasional harmful rewrites on abbreviations and biomedical identifiers. The safety gates provide a computationally inexpensive safeguard that preserves accuracy on Clean inputs with negligible overhead.

## VIII. DISCUSSION

*A. Interpretability and audit trails*

Because every correction decision is generated by explicit deterministic rules and accompanied by logged scores, the layer is inherently auditable: practitioners can trace each edit to its candidate list, score margin, and gate activations, thereby enabling rigorous post-hoc inspection, reproducible debugging, and closed-loop iterative refinement of conservative rules.

*B. Do-no-harm behavior*

The Safety run operationalizes the do-no-harm constraint as a measurable empirical quantity. Passing clean text through the layer yields near-clean macro-F1 with only a minimal drop, consistent with the intrinsic finding of zero harmful edits on negative controls and confirming that the conservative policy achieves its primary safety objective.

*C. Threats to validity*

Our downstream labels are derived from weak supervision and are therefore imperfect; however, the objective is robustness comparison under a fixed labeling policy rather than absolute performance optimization. The synthetic corruption model approximates OCR noise but cannot fully capture all real-world PDF parser behaviors. Furthermore, while our synthetic corruption model effectively approximates OCR noise for controlled evaluation, it is important to acknowledge that real-world PDF parser artifacts, such as layout-induced token merges or complex table extraction errors, may not be fully captured. This limitation should be considered when contextualizing the generalizability of these results to highly complex document layouts. Finally, our primary gains are demonstrated on a lexical classifier; neural encoders may exhibit different sensitivity profiles, as investigated in our supplementary case study below.

*D. Limitations and future work*

A principal limitation is semantic shallowness on long-tail terminology: purely count-based contextual evidence may be sparse for rare biomedical entities. To address this, we propose a disciplined future transition from the current purely white-box reliability layer toward a grey-box architecture that retains full auditability and safety guarantees while strategically leveraging neural components as bounded assistants.

In this proposed framework, the deterministic core; edit-distance candidate generation, corpus-derived n-gram scoring, explicit margin-based abstention, and biomedical safety gates, remains the primary decision engine. Neural signals would be admitted exclusively through a transparent "neural-assist gate" activated only under well-defined uncertainty conditions: low score margins ($<\delta$), real-word lexical ambiguities, entity-risk patterns (gene symbols, drug-like strings, alphanumeric identifiers), or out-of-distribution tokens. When triggered, the system would solicit narrow, logged, and interpretable auxiliary features, including: (i) POS compatibility constraints derived from a lightweight biomedical tagger; (ii) scalar semantic plausibility differentials computed from contextual embeddings (e.g., BioBERT sentence representations); and (iii) protective shields from domain lexicons such as UMLS that automatically tighten abstention thresholds or designate linked entity spans as immutable. This hybrid formulation preserves the asymmetric-risk philosophy of biomedical NLP, prioritizing abstention over potentially harmful edits, while mitigating the limitations of shallow context models on long-tail terminology, thereby offering a reproducible pathway for robust preprocessing in high-stakes clinical pipelines.

*TABLE VI:*

ERROR BUCKET TAXONOMY FOR AUDITING AND IMPROVEMENT.

| Bucket | Typical cause / action |
|---|---|
| Hyphenation fragments | repair split words when context is strong |
| Single-char substitution | common OCR confusions (O/0, l/1) |
| Merges/splits | conservative fixes; often abstain |

### E. *Performance with Neural Encoders (Case Study)*

To rigorously evaluate the downstream utility of the proposed reliability layer in contemporary neural pipelines, we conducted a targeted case study using 103 abstracts sampled from the official CORD-19 pre-parsed JSON corpus. Documents were restricted to a maximum of 500 tokens to respect BioBERT's input length constraints. Ground-truth topic labels (Prevention, Treatment, Epidemiology) were manually annotated by the authors to enable precise macro-F1 computation, independent of the weak-supervision scheme used in the main experiments. Each abstract was processed under three controlled conditions: (i) raw parsed input, (ii) input restored solely by the conservative reliability layer, and (iii) further normalized using scispaCy [13] with UMLS entity linking to preferred concept terms. All documents were then classified using a fine-tuned BioBERT encoder on the same tri-class topic task.

Contrary to expectations under purely lexical models, macro-F1 scores exhibited only marginal or statistically indistinguishable improvements across the three pipelines. This null result reveals a critical insight into the inductive biases of transformer-based encoders: owing to subword tokenization (WordPiece) and large-scale pre-training on noisy biomedical text, BioBERT possesses substantial intrinsic robustness to mild extraction artifacts. Orthographic distortions such as hyphenation fragments, single-character substitutions, or minor merges are routinely decomposed into semantically coherent sub-units whose meaning is recoverable from rich contextual representations, rendering explicit spell correction largely cosmetic rather than functionally transformative for the downstream classifier in this specific setting.

Although the modest sample size (n=103) limits statistical power relative to the main lexical evaluation (n=10,000), the null result is consistent with recent findings on the intrinsic robustness of transformer-based models to mild surface-level extraction noise. This suggests that the reliability layer provides the greatest benefit in lexical or hybrid pipelines, while offering a lightweight "sanity check" preprocessing step even in fully neural settings.

### F. *Generalization across corpora*

Biomedical subdomains differ in terminology. Because our scoring uses corpus-derived priors, artifacts should be rebuilt for each target corpus rather than reused blindly. The audit UI helps reveal when the vocabulary prior is mismatched (e.g., systematic abstention on valid terms), signaling the need for domain-specific artifacts or external lexicon support.

### G. *Safety considerations*

This work does not make clinical decisions; it is a preprocessing component intended to reduce brittleness of text pipelines. Nevertheless, the design explicitly assumes that harmful edits are costlier than missed fixes. The Safety run operationalizes this assumption as a measurable constraint, and abstention provides a clear safe fallback.

## IX. CONCLUSION

We presented a conservative, fully auditable spell-correction reliability layer for biomedical topic classification. By employing explicit margin-based abstention and biomedical safety gates, the layer recovers approximately 80.45% of the performance degradation induced by OCR-like noise while preserving near-clean behavior on pristine inputs, with zero harmful edits on negative controls. The system is deterministic and artifact-driven, with a clear boundary between learned artifacts and runtime logic that ensures exact reproducibility and supports practical deployment via edit-rate monitoring, raw-text retention, and artifact versioning.

A supplementary BioBERT case study suggests relative robustness to mild noise and motivates a principled grey-box roadmap integrating UMLS lexicons and bounded neural signals without compromising auditability. The approach is reproducible, deployable, and well-suited for robust biomedical NLP pipelines that demand transparency and safety.

For full reproducibility, the source code, persistent artifacts, and all experimental scripts are openly available at: https://github.com/MoustafaMohamedMoustafaHassan/the_signal_in_the_noise_CORD19_project (MIT license).

## X. REFERENCES

[1] L. L. Wang et al., "CORD-19: The COVID-19 Open Research Dataset," arXiv:2004.10706, 2020.

[2] K. Lo et al., "S2ORC: The Semantic Scholar Open Research Corpus," arXiv:1911.02782, 2020.

[3] F. J. Damerau, "A technique for computer detection and correction of spelling errors," Communications of the ACM, 1964.

[4] V. I. Levenshtein, "Binary codes capable of correcting deletions, insertions and reversals," Soviet Physics Doklady, 1966.

[5] W. Garbe, "SymSpell: Symmetric Delete Spelling Correction," project repository, accessed 2026.

[6] M. D. Kernighan, K. W. Church, and W. A. Gale, "A spelling correction program based on a noisy channel model," COLING, 1990.

[7] A. M. Cohen and W. R. Hersh, "A survey of current work in biomedical text mining," Briefings in Bioinformatics, 2005.

[8] O. Bodenreider, "The Unified Medical Language System (UMLS): integrating biomedical terminology," Nucleic Acids Research, 2004.

[9] M. Belinkov and Y. Bisk, "Synthetic and Natural Noise Both Break Neural Machine Translation," arXiv:1711.02173, 2017.

[10] A. Ratner et al., "Snorkel: Rapid Training Data Creation with Weak Supervision," VLDB, 2017.

[11] J. Wiens et al., "Do no harm: a roadmap for responsible machine learning for health care," Nature Medicine, 2019.

[12] J. Lee et al., "BioBERT: a pre-trained biomedical language representation model for biomedical text mining," Bioinformatics, 2020.

[13] M. Neumann et al., "ScispaCy: Fast and Robust Models for Biomedical Natural Language Processing," ACL, 2019.

[14] D. Pruthi, B. Dhingra, and Z. C. Lipton, “Combating Adversarial Misspellings with Robust Word Recognition,” arXiv:1905.11